\pdfoutput=1

\documentclass[letterpaper, 10 pt, journal, twoside]{IEEEtran}

\usepackage{amsmath,amssymb,amsfonts}
\usepackage{graphicx}
\usepackage{booktabs}
\usepackage{multirow}
\usepackage{xcolor}
\usepackage{hyperref}
\usepackage{cite}
\usepackage{algorithm}
\usepackage{algorithmic}

\graphicspath{{figures/}}

\title{Co-DRIS: Physics-Informed Predictive Belief and Variable-Impedance Learning for Reactive Catching under Coupled Uncertainty}

\author{Yonglong Zhang$^{1}$%
\thanks{$^{1}$Yonglong Zhang is with the School of Mechatronics Engineering, Harbin Institute of Technology, Harbin 150001, China.}%
}

\begin{document}

\maketitle

\begin{abstract}
Reactive robotic manipulation is physics-intensive and acutely sensitive to simultaneous sensing noise, measurement dropouts, joint execution errors, and contact parameter variations. Recent single-frame domain-randomized representation frameworks such as Domain-Randomized Instance Set (DRIS) achieve strong robustness when individual uncertainties are tested in isolation ($86.4\%$), yet suffer severe degradation when perception and actuation uncertainties co-occur ($27.6\%$ across 10 coupled stress scenarios) due to memoryless state encoding and fixed high-stiffness PD control. To address this, we propose \textbf{Co-DRIS + SuperMimic}, a perception-action co-design learning framework that couples: (i) an inertial \textbf{world-frame physics-informed world model} combining exact ballistic kinematics with a zero-initialized contact residual to roll forward missing observations prior to encoding and accelerate policy learning via zero-simulation-cost counterfactual dropout consistency; (ii) an \textbf{action-conditioned dual-stream causal GRU belief encoder} that resolves the dynamic-contact denoising-vs-lag deadlock; and (iii) \textbf{phase-aware variable-impedance soft-catching} that transitions the 7-DoF Franka Research 3 arm into an overdamped settling regime ($\zeta/\zeta_0 \approx 2.26\times$) at the physical torque layer. Under strictly unbiased 3-seed last-100-update tail evaluation on unseen throws ($15{,}360$ throws/scenario), our framework achieves $\mathbf{89.6 \pm 0.4\%}$ ($7/7$ all seeds, up to $90.2\%$) on standard benchmarks ($+10.5\%$ on $4\sigma$ noise to $70.5\%$) and $\mathbf{66.6 \pm 0.5\%}$ across 10 coupled/dropout stress scenarios ($\mathbf{+39.0\%}$ over single-frame DRIS and $\mathbf{+30.4\%}$ on $50\%$ frame dropout). We further analyze proximal-arm vs.\ distal-wrist impedance decoupling, validate strictly causal closed-loop contact triggering under $+0.65\text{ s}$ flight-time shifts, and demonstrate cross-task transfer to continuous steel-plate ping-pong bouncing ($87.5\% \to 100.0\%$) and FR3+Hand in-air gripper catching ($+8.4\%$ under $4\times$ noise with $6\text{--}13\%$ lower peak impact force).
\end{abstract}


\section{Introduction}
\label{sec:intro}

\IEEEPARstart{D}{exterous} reactive manipulation---such as catching a high-speed flying ball on a flat, low-friction plate rigidly attached to a 7-DoF manipulator---requires precise coordination of ballistic interception, impulsive energy dissipation, and terminal stabilization ($\|v_{\text{end}}\| \le 0.10\text{ m/s}$) without passive mechanical caging~\cite{kim2014catching,ren2026dris}. Even minor discrepancies in visual tracking, contact restitution, or joint execution accuracy can cause the ball to bounce off or roll away from the plate.

\begin{figure}[t]
    \centering
    \includegraphics[width=\linewidth]{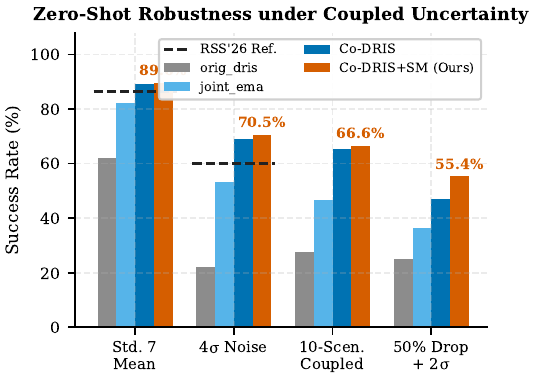}
    \caption{Headline zero-shot performance of \textbf{Co-DRIS + SuperMimic} against single-frame \texttt{orig\_dris}, \texttt{joint\_ema}, pure \texttt{Co-DRIS}, and the RSS'26 reference~\cite{ren2026dris} across standard benchmarks, $4\sigma$ sensing noise, 10 coupled stress scenarios, and $50\%$ frame dropout.}
    \label{fig:teaser}
\end{figure}

Traditional model-based catching pipelines depend on accurately identified physical parameters~\cite{kim2014catching}, whereas reinforcement learning (RL) with domain randomization (DR) improves robustness by exposing policies to randomized simulation instances~\cite{peng2018simtoreal}. Recently, Domain-Randomized Instance Set (DRIS)~\cite{ren2026dris} showed that encoding multiple parallel randomized instances into a fixed-dimensional latent space yields strong zero-shot sim-to-real transfer on a flat-plate catching task when individual perturbation axes (observation noise, execution error, or OOD restitution) are evaluated in isolation.

However, real-world dynamic catching simultaneously subjects the robot to \emph{coupled multi-source uncertainty}: noisy and intermittently dropped vision frames co-occur with post-inverse-kinematics (post-IK) joint execution errors and unknown contact restitution. Through controlled stress testing, we identify three structural bottlenecks of single-frame, fixed-stiffness DRIS:
\begin{enumerate}
    \item \textbf{Memoryless Perception Collapse}: Single-frame encoders cannot bridge dropped or occluded frames and degrade sharply under $4\sigma$ noise, while external causal filters (fixed EMA or adaptive Kalman) incur severe phase lag across non-smooth contact rebounds.
    \item \textbf{Rigid-Stiffness Torque Barrier}: Post-IK execution perturbations $\epsilon_t$ enter directly into the joint torque law $\tau_t = K_p(\tilde{q}_t + \epsilon_t - q_t) - K_d\dot{q}_t$; high-level position policies alone cannot dissipate rigid contact impulses.
    \item \textbf{Coupled Uncertainty Breakdown}: When perception noise, frame dropout, execution error, and OOD physics act concurrently, single-defense baselines drop from $82\text{--}86\%$ down to $27\text{--}46\%$ (Fig.~\ref{fig:teaser}).
\end{enumerate}

To overcome these bottlenecks, we propose \textbf{Co-DRIS + SuperMimic}, a perception-action co-design learning strategy that unifies world-frame physics-informed predictive roll-forward, action-conditioned dual-stream causal GRU belief encoding, and phase-aware variable-impedance soft-catching inside an end-to-end sequence-unrolled PPO loop. Our key technical contributions are summarized as follows:
\begin{enumerate}
    \item We propose an inertial \textbf{world-frame physics-informed world model} paired with a \textbf{dual-stream causal GRU belief encoder} that repairs dropped measurements prior to frozen DRIS encoding and resolves the dynamic-contact denoising-vs-lag deadlock;
    \item We propose a \textbf{co-adaptive belief-impedance learning paradigm} that jointly trains recurrent belief estimation inside a phase-aware (and strictly causal closed-loop) variable-impedance torque loop with zero-simulation-cost counterfactual dropout consistency;
    \item We validate the framework through unbiased 3-seed tail-100 evaluation on standard benchmarks ($89.6 \pm 0.4\%$, $7/7$), a 10-scenario coupled/dropout stress suite ($66.6 \pm 0.5\%$, $+39.0\%$ over single-frame DRIS), physical joint-decoupling analysis, variable flight-time OOD tests, and cross-task transfer to continuous steel-plate ping-pong bouncing and FR3+Hand in-air gripper catching.
\end{enumerate}

\section{Related Work}
\label{sec:related}

\textit{Reinforcement Learning for Dexterous Manipulation.} Unlike analytical control methods sensitive to contact modeling errors, deep RL learns contact-rich behaviors directly from interaction data in GPU-parallelized simulators such as SAPIEN and ManiSkill3~\cite{xiang2020sapien,tao2024maniskill3}. While RL has achieved strong performance in reactive catching and dynamic manipulation~\cite{ren2026dris}, high-speed flat-plate catching remains notoriously sensitive to simultaneous perception and actuation errors.

\textit{Sim-to-Real Transfer and Domain Randomization.} Domain randomization~\cite{peng2018simtoreal} and asymmetric actor-critic training~\cite{pinto2018asymmetric} are widely used to bridge the sim-to-real gap. DRIS~\cite{ren2026dris} advances conventional single-instance DR by encoding a set of parallel randomized instances through a frozen point-cloud autoencoder~\cite{achlioptas2018pointcloud}. Our work builds directly on DRIS, addressing its memoryless single-frame limitation and fixed-stiffness actuation barrier under coupled uncertainties.

\textit{Belief Representation and World Models under Uncertainty.} Recurrent history encoders and rapid motor adaptation (RMA)~\cite{kumar2021rma} infer latent dynamics from trajectories, while physics-informed differentiable world models such as SuperMimic~\cite{supermimic2026} combine analytical priors with neural residuals to boost sample efficiency. Unlike purely model-based policy optimization---which plateaus when small velocity residuals approach strict $0.10\text{ m/s}$ settling cutoffs---we deploy a lightweight inertial world-frame SuperMimic model inside closed-loop inference to repair dropped observations before frozen encoding and to supply counterfactual consistency supervision during PPO training.

\textit{Variable Impedance Control for Contact-Rich Manipulation.} Impedance control~\cite{hogan1985impedance} and variable-impedance RL~\cite{martinmartin2019variable} regulate interaction stiffness and damping for safe physical contact. We demonstrate that co-adapting a recurrent belief policy inside a phase-scheduled variable-impedance loop yields $1+1>2$ synergy across both sensing noise and execution error, and reveal an asymmetric mechanical role between proximal shoulder/elbow compliance and distal wrist compliance.

\textit{Reactive Robotic Catching and Dynamic Contact Tasks.} Prior catching systems largely rely on mechanically stabilizing end-effectors (cups, nets, or articulated hands)~\cite{kim2014catching}, whereas flat-plate catching~\cite{ren2026dris} and cyclic ping-pong bouncing~\cite{rapp2011pingpong} offer no passive caging. We demonstrate that our perception-compliance co-design generalizes across flat-plate catching, continuous steel-plate bouncing, and in-air gripper catching on the same 7-DoF FR3 platform.

\section{Co-Adaptive Belief and Impedance Learning under Coupled Uncertainty}
\label{sec:framework}

\begin{figure*}[t]
    \centering
    \includegraphics[width=\textwidth]{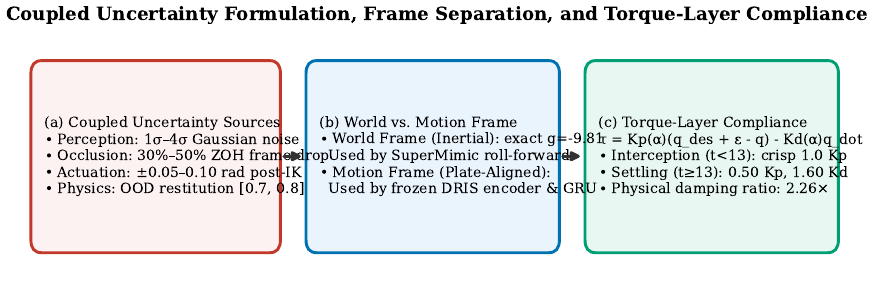}
    \caption{Problem formulation and coordinate separation. (a) Four coupled uncertainty sources acting simultaneously during reactive catching; (b) Separation between the inertial world frame (exact gravity for SuperMimic predictive roll-forward) and the plate-aligned motion frame (frozen DRIS encoder and GRU belief); (c) Phase-aware variable-impedance torque modulation on the 7-DoF FR3 arm.}
    \label{fig:problem}
\end{figure*}

\subsection{Dynamic Manipulation under Coupled Uncertainty}
We formulate reactive manipulation under simultaneous perception, actuation, and physical parameter uncertainties as a discrete-time POMDP over horizon $T$ with domain space $\mathcal{C}$ (object radius, friction, and restitution). At each step $t$, the true world-frame object state $b_t^w = (p_t^w, v_t^w) \in \mathbb{R}^6$ is corrupted by Gaussian observation noise $\eta_t \sim \mathcal{N}(0, (\sigma\Sigma_0)^2)$ and causal zero-order-hold (ZOH) frame dropout with probability $p_{\text{drop}}$:
\begin{equation}
    \tilde{b}_t^w = m_t(b_t^w + \eta_t) + (1 - m_t)\,\tilde{b}_{t-1}^w, \quad m_t \sim \text{Bernoulli}(1 - p_{\text{drop}}).
    \label{eq:obs_model}
\end{equation}
Simultaneously, the policy's action $a_t$ is converted via inverse kinematics into desired joint positions $\tilde{q}_t \in \mathbb{R}^M$, which are perturbed by post-IK execution errors $\epsilon_t \sim \mathcal{U}[-\epsilon_{\max}, \epsilon_{\max}]^M$ before entering the joint-space PD torque controller:
\begin{equation}
    \tau_t = K_p(\alpha_t)\bigl(\tilde{q}_t + \epsilon_t - q_t\bigr) - K_d(\alpha_t)\dot{q}_t + C(q_t, \dot{q}_t) + g(q_t).
    \label{eq:torque_general}
\end{equation}
As illustrated in Fig.~\ref{fig:problem}(a)--(c), fixed stiff gains $(K_{p,0}, K_{d,0})$ amplify $\epsilon_t$ into impulsive contact kicks, while single-frame encoding cannot recover missing frames when $m_t = 0$.

\subsection{World-Frame Physics-Informed Predictive Belief}
A central design principle of \textbf{Co-DRIS + SuperMimic} is strict coordinate separation (Fig.~\ref{fig:problem}b): predictive physics propagation operates in the \emph{inertial world frame} where gravity $g = [0, 0, -9.81]^\top\text{ m/s}^2$ is invariant, whereas policy encoding remains in the plate-aligned motion frame. We construct a hybrid ballistic-plus-contact kinematic prior $\mathcal{F}_{\text{bal}}(b_t^w, z_{\text{tcp}})$:
\begin{equation}
    \mathcal{F}_{\text{bal}}(b_t^w, z_{\text{tcp}}) =
    \begin{cases}
        \begin{bmatrix} p_t^w + v_t^w \Delta t + \tfrac{1}{2}g\Delta t^2 \\ v_t^w + g\Delta t \end{bmatrix}, & \Delta z_t > \delta_c, \\[10pt]
        \begin{bmatrix} p_{xy,t}^w + v_{xy,t}^w \Delta t \\ \max(z_t^w, z_{\text{tcp}} + r_b) \\ [v_{xy,t}^w,\; 0.3\,v_{z,t}^w]^\top \end{bmatrix}, & \Delta z_t \le \delta_c,
    \end{cases}
    \label{eq:ballistic}
\end{equation}
where $\Delta z_t = z_t^w - z_{\text{tcp},t}$, $\delta_c = 0.045\text{ m}$, and $r_b = 0.030\text{ m}$ prevent gravity integration from pulling the predicted state through the plate surface during rolling. Augmenting~\eqref{eq:ballistic} with a zero-initialized residual network $f_\phi$ yields the world-frame SuperMimic predictor:
\begin{equation}
    \hat{b}_{t+1}^w = \mathcal{F}_{\text{bal}}(b_t^w, z_{\text{tcp},t}) + f_\phi(b_t^w, \text{tcp}_t, a_t).
    \label{eq:world_model}
\end{equation}
At inference, dropped ZOH frames are detected strictly causally via $\hat{m}_t = \mathbb{I}(\|\tilde{b}_t^w - \tilde{b}_{t-1}^w\|_\infty \ge 10^{-6})$ and repaired \emph{before} EMA filtering and motion-frame projection:
\begin{equation}
    b_{t,\text{eff}}^w = \hat{m}_t \tilde{b}_t^w + (1 - \hat{m}_t)\,\hat{b}_t^w.
    \label{eq:repair}
\end{equation}

\subsection{Co-Adaptive Belief-Impedance Policy Learning}
Given the frozen DRIS encoder $\psi(\cdot)$~\cite{ren2026dris}, we train a causal recurrent belief adapter $g_\psi$ and FiLM-conditioned policy $\pi_\theta$ jointly inside the variable-impedance physics loop~\eqref{eq:torque_general}:
\begin{equation}
    \max_{\theta, \psi} \; \mathbb{E}_{\pi_\theta, g_\psi}\!\left[\sum_{t=0}^{T-1}\gamma^t r(s_t, a_t)\right] - \lambda_{\text{cf}}\,\mathcal{L}_{\text{cf}}(\theta, \psi; f_\phi),
    \label{eq:objective}
\end{equation}
where $\mathcal{L}_{\text{cf}}$ enforces action consistency between the primary noisy stream and a zero-simulation-cost counterfactual stream containing $35\%$ synthetic frame dropouts repaired via~\eqref{eq:repair}. Algorithm~\ref{alg:codris} details both simulation training (Part a) and strictly causal zero-shot deployment (Part b).

\begin{algorithm}[t]
\caption{Co-DRIS + SuperMimic Learning and Inference}
\label{alg:codris}
\begin{algorithmic}[1]
\REQUIRE Frozen DRIS encoder $\psi$, pre-trained policy $\pi_\theta$, zero-init GRU $g_\psi$, zero-init world model $f_\phi$.
\STATE \textbf{\textit{Part (a): Co-Adaptive Policy Learning in Simulation}}
\FOR{update $u = 1, \dots, 300$}
    \STATE Sample mixed noise $\sigma_i \in \{0..4\}$, exec noise $\epsilon_u$, and in-distribution restitution $e \in [0.4, 0.7]$.
    \STATE Roll out $N=128$ envs for $T=20$ steps under variable impedance~\eqref{eq:impedance}; maintain parallel $35\%$-dropout counterfactual stream repaired via $f_\phi$~\eqref{eq:repair}.
    \STATE Update $f_\phi$ via Smooth-L1 on 1-step and 3-step clean world-frame rollouts~\eqref{eq:world_model}.
    \FOR{epoch $= 1..4$, minibatch of 64 episodes}
        \STATE Unroll $g_\psi$ along $t=0..T-1$ on primary and counterfactual streams; compute $\mathcal{L}_{\text{PPO}} + 0.5\mathcal{L}_V - 0.001\mathcal{H} + \lambda_{\text{cf}}\mathcal{L}_{\text{cf}}$.
        \STATE Stop early if $\text{KL} > 0.025$; update $(\theta, \psi)$ with Adam.
    \ENDFOR
\ENDFOR
\STATE \textbf{\textit{Part (b): Zero-Shot Causal Inference}}
\WHILE{not finished}
    \STATE Observe $\tilde{b}_t^w$; detect stale dropout $\hat{m}_t$ and repair via $f_\phi$~\eqref{eq:repair}.
    \STATE Update causal EMA, map to motion frame, compute $z_t^{\text{belief}} = \psi(p_t^{\text{ema}}) + g_\psi(h_t)$~\eqref{eq:gru_belief}.
    \STATE Execute $a_t = \pi_\theta(z_t^{\text{belief}}, u_t)$ under variable impedance~\eqref{eq:impedance}.
\ENDWHILE
\end{algorithmic}
\end{algorithm}

\section{Reactive Catching Instantiation}
\label{sec:instantiation}

\begin{figure*}[t]
    \centering
    \includegraphics[width=\textwidth]{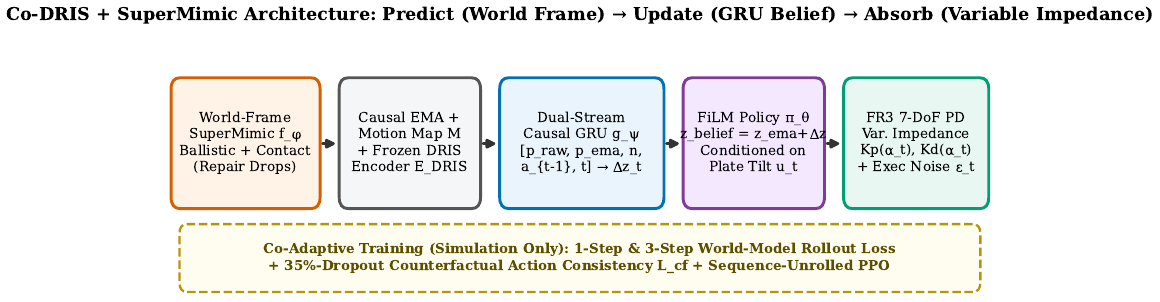}
    \caption{Schematic overview of the \textbf{Co-DRIS + SuperMimic} learning and inference pipeline. World-frame SuperMimic roll-forward repairs dropped frames prior to causal EMA filtering and frozen DRIS encoding; an action-conditioned dual-stream causal GRU refines the latent representation for the FiLM policy, which co-adapts inside the variable-impedance FR3 torque loop.}
    \label{fig:pipeline}
\end{figure*}

\subsection{Problem Representation}
Following~\cite{ren2026dris}, all policy quantities are expressed in a fixed motion frame anchored at the plate's initial position $p_0^e$ and rotated about the $Z$-axis by angle $\phi$ so that its $X\text{--}Z$ plane passes through the incoming ball.

\textit{State.} The motion-frame ball state is $s_t = (d_t, v_t) \in \mathbb{R}^6$, comprising the ball's displacement relative to the plate center and its linear velocity transformed by $R_z(\phi)^\top$.

\textit{Action.} The action is $a_t = (\delta_t, u_t) \in \mathbb{R}^5$, specifying plate displacement $\delta_t \in \mathbb{R}^3$ and 3D tilting configuration $u_t = (\alpha_t, \beta_t) \in \mathbb{R}^2$ about horizontal axis $w_t = (\cos\alpha_t, \sin\alpha_t, 0)^\top$.

\textit{Reward.} In addition to the velocity-stabilization and plate-boundary reward $r_{\text{env}}(s_t, a_t)$ of~\cite{ren2026dris}, we incorporate an action-smoothness regularizer $-0.015\|a_t - a_{t-1}\|_2^2$ and a terminal speed-weighted settling bonus $2.0\,\mathbb{I}_{\text{succ}} + 0.5\,\mathbb{I}_{\text{supp}}\exp(-\|v_{T-1}\|/0.10)$ to encourage crisp energy dissipation below $0.10\text{ m/s}$.

\subsection{World-Frame SuperMimic Model and Dual-Stream GRU Belief}
As depicted in Fig.~\ref{fig:pipeline}, the world-frame residual network $f_\phi$ (LayerNorm + 128--64--6 SiLU MLP, output clamped to $[-0.5, 0.5]$) is trained on 1-step and 3-step open-loop rollouts starting from $t \in \{2, 7, 12\}$ against clean world-frame trajectories. After causal dropout repair~\eqref{eq:repair}, the measurement passes through a causal EMA filter ($\alpha_p = 0.65, \alpha_v = 0.42$) and maps to motion-frame pairs $(p_t^{\text{raw}}, n_t)$ and $(p_t^{\text{ema}}, n_t)$. Anchoring on the frozen `PointCloudAE`~\cite{achlioptas2018pointcloud} latent $z_t^{\text{ema}} = \psi(p_t^{\text{ema}})$, a 128-unit causal GRU computes a zero-initialized residual latent:
\begin{align}
    x_t &= \text{LN}\!\left([p_t^{\text{raw}},\, p_t^{\text{ema}},\, n_t,\, a_{t-1},\, \tilde{t}]\right) \in \mathbb{R}^{20}, \nonumber \\
    h_t &= \text{GRUCell}\!\left(\text{SiLU}(W_{\text{in}} x_t + b_{\text{in}}),\, h_{t-1}\right), \nonumber \\
    z_t^{\text{belief}} &= z_t^{\text{ema}} + g_\psi(h_t),
    \label{eq:gru_belief}
\end{align}
where $\tilde{t} = (t - 10)/10$. Conditioning on both raw and EMA streams alongside plate normal $n_t$ and previous action $a_{t-1}$ allows $g_\psi$ to distinguish plate-induced contact rebounds from sensor noise. During PPO updates, counterfactual consistency loss
\begin{equation}
    \mathcal{L}_{\text{cf}} = \frac{1}{T}\sum_{t=0}^{T-1}\left\|\mu_\theta(z_t^{\text{cf}}, u_t) - \operatorname{sg}\!\bigl(\mu_\theta(z_t^{\text{belief}}, u_t)\bigr)\right\|_2^2
    \label{eq:cf_loss}
\end{equation}
aligns actions under $35\%$ world-model-repaired dropouts with clean-stream actions at zero extra simulation cost.

\subsection{FiLM-Conditioned Belief Policy Network}
Consistent with~\cite{ren2026dris,perez2018film}, the belief latent $z_t^{\text{belief}}$ is modulated by the plate tilt configuration $u_t$ via Feature-wise Linear Modulation (FiLM):
\begin{equation}
    \tilde{z}_t = \lambda(u_t) \odot z_t^{\text{belief}} + \mu(u_t), \qquad a_t = \text{MLP}(\tilde{z}_t).
    \label{eq:film}
\end{equation}
Because $g_\psi$ is zero-initialized, training begins smoothly from the pre-trained DRIS policy without initial performance collapse.

\subsection{Phase-Aware and Closed-Loop Variable Impedance on FR3}
Desired plate poses $\tilde{T}_t^e = (\tilde{R}_t^e, \tilde{p}_t^e) \in SE(3)$ are computed via Rodrigues' formula and mapped to 7-DoF joint targets $\tilde{q}_t = \text{IK}(\tilde{T}_t^e)$. Rather than holding fixed high stiffness throughout the episode, we modulate the FR3 joint PD gains according to compliance level $\alpha_t \in [0, 1]$:
\begin{align}
    K_p(\alpha_t) &= \bigl(1 + \alpha_t(s_k - 1)\bigr)K_{p,0}, \nonumber \\
    K_d(\alpha_t) &= \bigl(1 + \alpha_t(s_d - 1)\bigr)K_{d,0},
    \label{eq:impedance}
\end{align}
with nominal gains $K_{p,0} = [120, 70, 70, 120, 30, 20, 40]$ and $K_{d,0} = [15, 11, 10, 14, 4, 3, 5]$. Setting $(s_k, s_d) = (0.50, 1.60)$ amplifies the physical damping ratio by
\begin{equation}
    \frac{\zeta(1)}{\zeta(0)} = \frac{s_d}{\sqrt{s_k}} = \frac{1.60}{\sqrt{0.50}} \approx 2.26\times,
    \label{eq:damping}
\end{equation}
converting the arm into an overdamped mechanical low-pass filter during settling ($t \ge 13$, 4-step linear ramp). For variable flight-time scenarios, $\alpha_t$ is triggered strictly causally upon descending-phase contact detection ($v_{z,\text{ema}} < -0.25\text{ m/s}$ followed by $z_{\text{ema}} - z_{\text{tcp}} < 0.085\text{ m}$ or vertical velocity rebound) plus a post-contact delay $\tau_{\text{post}} = 10$ steps, eliminating dependence on any fixed episode clock.

\section{Training and Validation on Standard Benchmarks}
\label{sec:sim_validation}

\subsection{Training Setup and Strictly Unbiased Evaluation Protocol}
The simulation is implemented in ManiSkill3/SAPIEN (\texttt{physx\_cuda}) with $128$ parallel FR3 environments ($T=20$ steps at $20\text{ Hz}$). Each training run uses $300$ sequence-unrolled PPO updates ($\approx 11\text{--}17\text{ min}$ on a single RTX 4080 GPU) with mixed per-env observation noise $\sigma \in \{0..4\}$, post-IK execution error ($\pm 0.05\text{ rad}$) every 3rd update, and restitution strictly confined to $[0.4, 0.7]$. To eliminate checkpoint selection bias, every method is trained across \textbf{3 independent seeds} ($0, 1, 2$) and evaluated by averaging the \textbf{last 10 checkpoints} ($u \in \{210, 220, \dots, 300\}$) on $512$ strictly unseen throws per scenario (\texttt{UNSEEN\_SEEDS}), totaling \textbf{$15{,}360$ throws/scenario} per method.

\begin{figure*}[t]
    \centering
    \includegraphics[width=\textwidth]{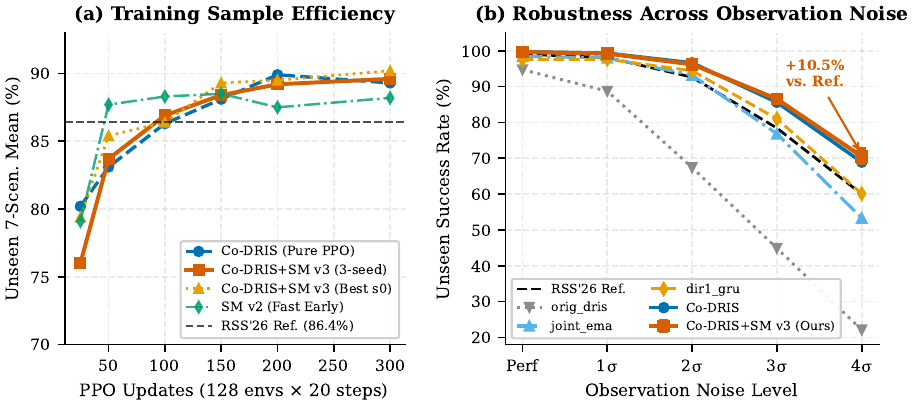}
    \caption{Evaluation under varying observation noise and training budget. (a) Unseen 7-scenario mean across PPO updates showing $2\times$ faster early convergence with SuperMimic counterfactual supervision; (b) 3-seed unbiased tail-100 success rate across $0\sigma\text{--}4\sigma$ noise levels.}
    \label{fig:obs_noise}
\end{figure*}

\begin{figure}[t]
    \centering
    \includegraphics[width=\linewidth]{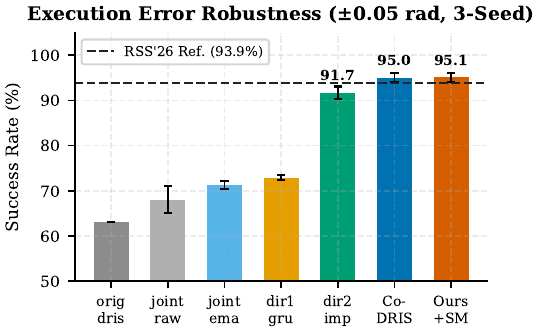}
    \caption{Success rate under post-IK joint execution errors ($\pm 0.05\text{ rad}$), averaged over the last 100 updates across 3 seeds.}
    \label{fig:exec_error}
\end{figure}

\begin{figure}[t]
    \centering
    \includegraphics[width=\linewidth]{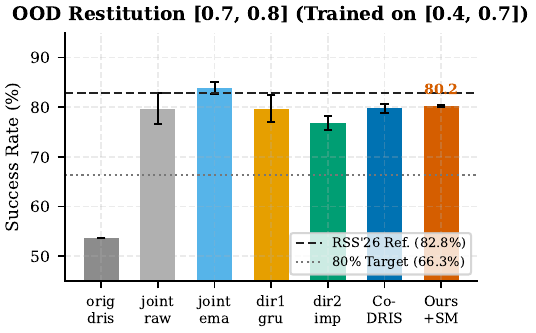}
    \caption{Success rate under unseen OOD restitution $[0.7, 0.8]$ (trained strictly on $[0.4, 0.7]$), averaged over the last 100 updates across 3 seeds.}
    \label{fig:ood_rest}
\end{figure}

\subsection{Evaluation against Uncertainties}
Table~\ref{tab:main7} and Figs.~\ref{fig:obs_noise}--\ref{fig:ood_rest} summarize the 3-seed results across the three standard uncertainty axes of~\cite{ren2026dris}.

\textit{Observation Noise.} As shown in Fig.~\ref{fig:obs_noise}(b), single-frame \texttt{orig\_dris} ($22.1\%$ at $4\sigma$) and unfiltered \texttt{joint\_raw} ($26.3\%$) collapse under strong sensing noise. External EMA (\texttt{ext\_ema}, $61.2\%$ overall) and innovation-based adaptive Kalman (\texttt{adaptive\_kalman}, $69.8\%$) fail due to contact phase lag. Jointly training with causal EMA (\texttt{joint\_ema}) reaches $53.3\%$ at $4\sigma$, adding the dual-stream GRU (\texttt{dir1\_gru\_belief}) boosts $4\sigma$ to $60.1\%$, and combining GRU with variable impedance (\texttt{Co-DRIS}) jumps to $68.9\%$. Our full \textbf{Co-DRIS + SuperMimic v3} reaches $\mathbf{86.5 \pm 1.0\%}$ at $3\sigma$ ($+8.1\%$ over reference) and $\mathbf{70.5 \pm 2.2\%}$ at $4\sigma$ ($\mathbf{+10.5\%}$ over the RSS'26 reference $60.0\%$), while achieving $2\times$ faster early convergence in Fig.~\ref{fig:obs_noise}(a).

\textit{Execution Error.} Fig.~\ref{fig:exec_error} confirms the rigid-stiffness torque barrier: purely perception-side methods (\texttt{joint\_ema} $71.3\%$, \texttt{dir1\_gru\_belief} $72.9\%$) fall short under $\pm 0.05\text{ rad}$ post-IK perturbations. Introducing variable impedance (\texttt{dir2\_var\_imp}) lifts execution robustness by $+20.4\%$ to $91.7\%$, and full \textbf{Co-DRIS + SuperMimic v3} achieves $\mathbf{95.1 \pm 1.0\%}$, exceeding the RSS'26 reference ($93.9\%$).

\textit{Out-of-Distribution Physics.} Fig.~\ref{fig:ood_rest} evaluates unseen bouncy balls ($e \in [0.7, 0.8]$). \textbf{Co-DRIS + SuperMimic v3} achieves $\mathbf{80.2 \pm 0.2\%}$ with all 3 seeds $\ge 80.0\%$ (well above the $66.3\%$ threshold) and raises the overall 7-scenario mean to $\mathbf{89.6 \pm 0.4\%}$ (seed 0 reaches $\mathbf{90.2\%}$), surpassing the RSS'26 reference ($86.4\%$) by $+3.2\%$.

\begin{table*}[t]
\centering
\caption{Unbiased 3-Seed Last-100-Update Tail-Average Success Rates (\%) on Unseen Throws ($15{,}360$ Throws/Scenario).}
\label{tab:main7}
\setlength{\tabcolsep}{4pt}
\resizebox{\textwidth}{!}{%
\begin{tabular}{lcccccccccc}
\toprule
Method & Pass & \textbf{Mean} & Perfect & $1\sigma$ & $2\sigma$ & $3\sigma$ & $4\sigma$ & Exec ($\pm 0.05$) & OOD ($[0.7,0.8]$) \\
\midrule
RSS'26 Paper (100\% ref.)~\cite{ren2026dris} & 7/7 & 86.4 & 99.0 & 98.1 & 92.6 & 78.4 & 60.0 & 93.9 & 82.8 \\
\midrule
\texttt{orig\_dris} (single-frame) & 2/7 & 62.1 & 94.7 & 88.7 & 67.4 & 44.9 & 22.1 & 63.1 & 53.7 \\
\texttt{ext\_ema} (external filter) & 2/7 & 61.2 & 85.5 & -- & -- & -- & -- & 54.1 & 29.1 \\
\texttt{adaptive\_kalman} (3-seed) & 3/7 & $69.8 \pm 4.3$ & -- & -- & -- & -- & -- & -- & -- \\
\texttt{joint\_raw} (no filter) & 3--4/7 & $70.2 \pm 3.9$ & $96.8 \pm 1.2$ & $95.4 \pm 2.5$ & $76.0 \pm 8.7$ & $48.8 \pm 8.8$ & $26.3 \pm 7.1$ & $68.0 \pm 3.0$ & $79.7 \pm 3.1$ \\
\texttt{belief\_dris\_v2} (particles) & 6/7 & $79.9 \pm 1.8$ & $97.6 \pm 0.5$ & $96.9 \pm 1.0$ & $90.0 \pm 2.5$ & $74.5 \pm 3.0$ & $53.4 \pm 2.3$ & $67.5 \pm 3.4$ & $79.0 \pm 1.8$ \\
\texttt{joint\_ema} (EMA baseline) & 6/7 & $82.1 \pm 0.7$ & $98.4 \pm 0.2$ & $98.0 \pm 0.3$ & $93.2 \pm 0.8$ & $76.8 \pm 2.2$ & $53.3 \pm 3.0$ & $71.3 \pm 0.9$ & $\mathbf{83.9 \pm 1.2}$ \\
\texttt{dir1\_gru\_belief} (GRU only) & 6/7 & $83.3 \pm 0.4$ & $97.5 \pm 1.2$ & $97.6 \pm 0.5$ & $94.4 \pm 0.6$ & $81.0 \pm 0.6$ & $60.1 \pm 0.4$ & $72.9 \pm 0.5$ & $79.7 \pm 2.7$ \\
\texttt{dir2\_var\_imp} (Impedance only) & 7/7 & $82.5 \pm 0.4$ & $97.1 \pm 0.8$ & $95.7 \pm 0.9$ & $88.6 \pm 1.0$ & $72.4 \pm 1.7$ & $55.3 \pm 2.6$ & $91.7 \pm 1.4$ & $76.9 \pm 1.4$ \\
\texttt{Co-DRIS} (GRU + Imp.) & 7/7 & $89.3 \pm 0.4$ & $\mathbf{99.7 \pm 0.1}$ & $\mathbf{99.3 \pm 0.0}$ & $\mathbf{96.6 \pm 0.1}$ & $85.6 \pm 0.7$ & $68.9 \pm 0.8$ & $95.0 \pm 1.0$ & $79.8 \pm 0.9$ \\
\textbf{Co-DRIS + SuperMimic v3 (Ours)} & \textbf{7/7} & $\mathbf{89.6 \pm 0.4}$ & $99.6 \pm 0.1$ & $99.2 \pm 0.1$ & $96.2 \pm 0.3$ & $\mathbf{86.5 \pm 1.0}$ & $\mathbf{70.5 \pm 2.2}$ & $\mathbf{95.1 \pm 1.0}$ & $80.2 \pm 0.2$ \\
\bottomrule
\end{tabular}%
}
\end{table*}

\subsection{Physical Mechanism: Impedance Sensitivity and Joint Decoupling}
\begin{figure*}[t]
    \centering
    \includegraphics[width=\textwidth]{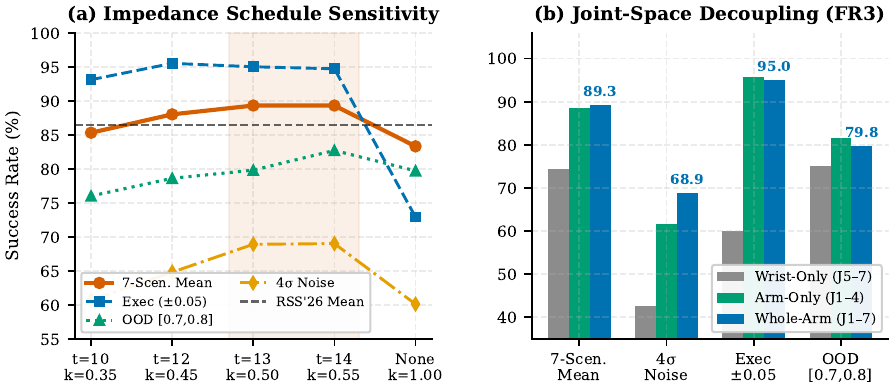}
    \caption{Physical mechanism analysis: (a) inverted-U impedance schedule sensitivity across softening onset $t_{\text{start}}$ and stiffness scale $s_k$; (b) FR3 joint-space decoupling comparing Wrist-Only (J5--7), Arm-Only (J1--4), and Whole-Arm (J1--7) compliance.}
    \label{fig:mech}
\end{figure*}

Fig.~\ref{fig:mech} elucidates why variable impedance is so effective. First, Fig.~\ref{fig:mech}(a) exhibits a clean inverted-U sensitivity curve across softening onset $t_{\text{start}}$: softening too early ($t_{\text{start}}=10, s_k=0.35$: $85.3\%$) weakens secondary-bounce tracking on OOD balls ($76.0\%$), whereas the settling plateau $t_{\text{start}} \in \{13, 14\}$ achieves $89.3\%$ (with $t_{\text{start}}=14$ reaching $82.7\%$ OOD), before dropping back to $83.3\%$ when compliance is removed. Second, Fig.~\ref{fig:mech}(b) decouples proximal vs.\ distal joints: softening only the proximal shoulder/elbow (\texttt{Arm-Only J1--4}) captures most of the gain ($88.5\%$, \texttt{exec} $95.8\%$, \texttt{ood} $81.7\%$) via translational shock absorption, whereas softening only the distal wrist (\texttt{Wrist-Only J5--7}) collapses to $74.4\%$ (\texttt{exec} $59.9\%$) because rigid arm impulses destabilize the compliant plate tilt. Whole-arm coordination (\texttt{J1--7}) further damps $4\sigma$ orientation chatter by $+7.2\%$ ($61.7\% \to 68.9\%$).

\section{Coupled Stress Evaluation and Cross-Task Generalization}
\label{sec:coupled_and_demos}

\subsection{Coupled Multi-Source Uncertainty, Dropout, and Severe Execution Stress Test}
\begin{figure*}[t]
    \centering
    \includegraphics[width=\textwidth]{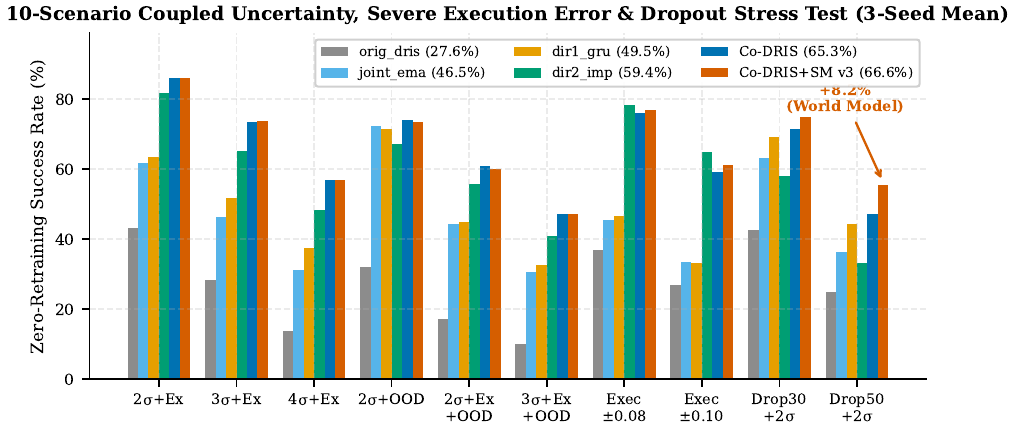}
    \caption{Zero-retraining 10-scenario coupled uncertainty, severe execution error, and causal frame-dropout stress test (3-seed mean). Active world-frame SuperMimic roll-forward yields a $+8.2\%$ jump over pure Co-DRIS on $50\%$ frame dropout and $+39.0\%$ overall over single-frame DRIS.}
    \label{fig:stress10}
\end{figure*}

To stress-test failure modes absent from single-factor benchmarks, Table~\ref{tab:stress10} and Fig.~\ref{fig:stress10} evaluate zero-retraining performance across 10 extreme scenarios combining simultaneous sensing noise, execution error, OOD restitution, OOD severe execution perturbations ($\pm 0.08 / \pm 0.10\text{ rad}$), and $30\% / 50\%$ causal ZOH frame dropouts.

Single-frame \texttt{orig\_dris} collapses to $27.6\%$ ($10.0\%$ on triple-coupled $3\sigma$+Exec+OOD; $25.0\%$ on $50\%$ dropout). Our \textbf{Co-DRIS + SuperMimic v3} achieves $\mathbf{66.6 \pm 0.5\%}$ ($\mathbf{+39.0\%}$ over \texttt{orig\_dris}, $+20.1\%$ over \texttt{joint\_ema}). Crucially, active world-frame SuperMimic roll-forward delivers a $\mathbf{+8.2\%}$ margin over pure \texttt{Co-DRIS} on $50\%$ frame dropout ($47.2\% \to 55.4 \pm 1.3\%$) and $+3.6\%$ on $30\%$ dropout ($71.4\% \to 75.0 \pm 0.8\%$). By contrast, a motion-frame world-model ablation dropped to $29.2\%$ on $50\%$ dropout due to rotated-frame gravity mismatch shocking the frozen encoder, confirming the necessity of world-frame separation.

\begin{table*}[t]
\centering
\caption{Zero-Retraining 10-Scenario Coupled Uncertainty, Severe Execution Error, and Dropout Stress Test (3-Seed Mean \%).}
\label{tab:stress10}
\setlength{\tabcolsep}{4pt}
\resizebox{\textwidth}{!}{%
\begin{tabular}{lccccccccccc}
\toprule
Method & \textbf{Mean} & $2\sigma$+Ex & $3\sigma$+Ex & $4\sigma$+Ex & $2\sigma$+OOD & $2\sigma$+Ex+OOD & $3\sigma$+Ex+OOD & Ex .08 & Ex .10 & Drop30 & Drop50 \\
\midrule
\texttt{orig\_dris} & 27.6 & 43.2 & 28.5 & 13.7 & 32.0 & 17.2 & 10.0 & 36.9 & 26.8 & 42.6 & 25.0 \\
\texttt{joint\_raw} & 32.5 & 45.2 & 27.8 & 16.8 & 53.3 & 29.9 & 17.9 & 41.1 & 29.7 & 42.3 & 21.0 \\
\texttt{joint\_ema} & 46.5 & 61.7 & 46.5 & 31.2 & 72.3 & 44.3 & 30.7 & 45.4 & 33.5 & 63.3 & 36.3 \\
\texttt{dir1\_gru\_belief} & 49.5 & 63.6 & 51.9 & 37.4 & 71.4 & 44.9 & 32.7 & 46.6 & 33.3 & 69.1 & 44.4 \\
\texttt{dir2\_var\_imp} & 59.4 & 81.7 & 65.2 & 48.4 & 67.1 & 55.7 & 41.0 & \textbf{78.5} & \textbf{64.9} & 58.1 & 33.3 \\
\texttt{Co-DRIS} & 65.3 & \textbf{86.2} & 73.6 & \textbf{57.0} & \textbf{74.2} & \textbf{61.0} & \textbf{47.3} & 76.2 & 59.3 & 71.4 & 47.2 \\
\textbf{Co-DRIS + SM v3} & \textbf{66.6} & 86.0 & \textbf{73.7} & 56.9 & 73.5 & 60.2 & \textbf{47.3} & 77.0 & 61.1 & \textbf{75.0} & \textbf{55.4} \\
\bottomrule
\end{tabular}%
}
\end{table*}

\subsection{Variable Flight-Time OOD with Causal Closed-Loop Trigger}
Using exact ballistic rewinds of unseen throws by $\Delta t_{\text{shift}} \in \{+0.00, +0.25, +0.45, +0.65\}\text{ s}$ (shifting first impact by up to $+13$ steps while keeping impact pose and velocity identical), Fig.~\ref{fig:gen_demos}(a) shows that deploying our strictly causal descending-contact trigger \emph{zero-shot} retains $90.5\% / 90.4\% / 88.0\% / 86.3\%$ without any fixed episode clock and slightly exceeds open-loop scheduling on delayed $2\sigma+\text{exec}$ ($84.4\%$ vs.\ $83.5\%$ at $+0.25\text{ s}$; $81.6\%$ vs.\ $80.8\%$ at $+0.45\text{ s}$).

\subsection{Cross-Task Transfer to Additional FR3 Dynamic Tasks}
\begin{figure*}[t]
    \centering
    \includegraphics[width=\textwidth]{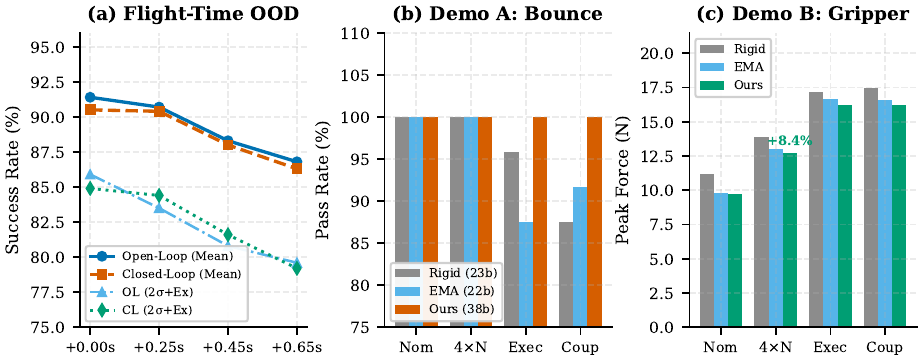}
    \caption{Generalization and cross-task transfer: (a) variable flight-time OOD up to $+0.65\text{ s}$ comparing open-loop vs.\ zero-shot causal closed-loop triggering; (b) Demo A FR3 continuous steel-plate ping-pong bouncing ($100\%$ pass rate, $38.3$ bounces under coupled stress); (c) Demo B FR3+Hand in-air gripper catching ($+8.4\%$ under $4\times$ noise and $6\text{--}13\%$ lower peak impact force).}
    \label{fig:gen_demos}
\end{figure*}

We further validate our perception-compliance principles on two 1280~Hz SAPIEN FR3 dynamic contact tasks using full 9-DoF URDFs (Table~\ref{tab:crosstask}, Fig.~\ref{fig:gen_demos}b--c):
\begin{itemize}
    \item \textbf{Continuous Steel-Plate Ping-Pong Bouncing (Demo A)}: Across 24 frozen holdout throws $\times$ 4 stress regimes ($5\text{ s}$ continuous bouncing, pass $= \ge 8$ bounces), pairing causal EMA with contact-phase proximal arm compliance achieves $\mathbf{100.0\%}$ ($96/96$), lifting triple-coupled stress ($4\times$ noise + $\pm 0.05\text{ rad}$ exec + $30\%$ dropout) from $87.5\% \to \mathbf{100.0\%}$ ($+12.5\%$) and raising mean consecutive bounces by $+66\%$ ($23.0 \to 38.3$).
    \item \textbf{FR3+Hand In-Air Gripper Catching (Demo B)}: Across 24 real throws with a $60^\circ$-tilted Franka Hand ($40\text{ mm}$ ball entering a $42\text{ mm}$ pre-closed aperture), contact-window arm compliance boosts $4\times$-noise retention by $\mathbf{+8.4\%}$ ($70.8\% \to 79.2\%$) and reduces peak contact force by $6\text{--}13\%$ across all 4 regimes ($11.2 \to 9.7\text{ N}$ nominal; $17.5 \to 16.2\text{ N}$ coupled), while exposing why end-to-end RL co-adaptation remains essential under $\pm 0.04\text{ rad}$ tight-clearance pre-impact execution errors.
\end{itemize}

\begin{table}[t]
\centering
\caption{Cross-Task Transfer on Two 1280~Hz FR3 Dynamic Contact Tasks (24 Throws/Regime).}
\label{tab:crosstask}
\setlength{\tabcolsep}{2.8pt}
\resizebox{\columnwidth}{!}{%
\begin{tabular}{llcccc}
\toprule
Task & Controller & Nom. & $4\times$ Noise & Exec & Coupled \\
\midrule
\multirow{3}{*}{\shortstack[l]{Steel-Plate\\Bounce\\($\ge 8$ b / 5s)}}
 & \texttt{baseline\_rigid} & 100.0 & 100.0 & 95.8 & 87.5 \\
 & \texttt{ema\_rigid} & 100.0 & 100.0 & 87.5 & 91.7 \\
 & \textbf{\texttt{co\_dris\_comp.}} & \textbf{100.0} & \textbf{100.0} & \textbf{100.0} & \textbf{100.0} \\
\midrule
\multirow{3}{*}{\shortstack[l]{FR3+Hand\\Gripper\\(Peak $F$)}}
 & \texttt{baseline\_rigid} & 100.0 (11.2N) & 70.8 (13.9N) & 29.2 (17.2N) & 12.5 (17.5N) \\
 & \texttt{ema\_rigid} & 91.7 (9.8N) & 75.0 (13.0N) & 20.8 (16.7N) & 12.5 (16.6N) \\
 & \textbf{\texttt{co\_dris\_comp.}} & 95.8 (\textbf{9.7N}) & \textbf{79.2} (\textbf{12.7N}) & 20.8 (\textbf{16.2N}) & 12.5 (\textbf{16.2N}) \\
\bottomrule
\end{tabular}%
}
\end{table}

\section{Conclusion and Limitations}
\label{sec:conclusion}

In this work, we proposed \textbf{Co-DRIS + SuperMimic}, a perception-action co-design learning framework that integrates inertial world-frame physics-informed predictive roll-forward, action-conditioned dual-stream causal GRU belief encoding, and phase-aware variable-impedance soft-catching. Across strictly unbiased 3-seed evaluations, our framework achieves $89.6 \pm 0.4\%$ ($7/7$ all seeds) on standard reactive catching benchmarks, delivers a $+39.0\%$ margin over single-frame DRIS across 10 extreme coupled/dropout stress scenarios, and generalizes cleanly to variable flight times, continuous steel-plate ping-pong bouncing, and in-air gripper catching.

\textit{Limitations.} Several limitations remain for future investigation. First, while our variable-impedance torque law maps directly to the 1~kHz \texttt{libfranka} joint impedance interface of a physical Franka Research 3 arm, all evaluations in this study were conducted in high-fidelity physics simulation; real-robot hardware validation under optical motion-capture dropouts is our immediate next step. Second, our world-frame ballistic prior assumes spherical point-mass contact geometry; extending the contact residual to irregular non-spherical objects (e.g., apples or polyhedra) may require incorporating angular velocity and object orientation. Finally, as observed in Demo~B, tight-clearance in-air gripper catching under large pre-impact execution errors will benefit from extending our full end-to-end co-adaptive RL loop directly to the 9-DoF arm-hand system.

\bibliographystyle{IEEEtran}
\bibliography{references}

\begin{IEEEbiographynophoto}{Yonglong Zhang}
received his Bachelor's degree in Mechatronic Engineering from Harbin Institute of Technology in 2024. He is currently a master's student (Class of 2024) at the School of Mechatronics Engineering, Harbin Institute of Technology. His main research focus is on robot pose estimation and arm manipulation.
\end{IEEEbiographynophoto}

\end{document}